%% file: main.tex
\pdfoutput=1
\documentclass{article}

\usepackage{microtype}
\usepackage{graphicx}
\usepackage{subfigure}
\usepackage{booktabs} %

\usepackage{hyperref}

\usepackage[accepted]{icml2025}

\usepackage{amsmath}
\usepackage{amssymb}
\usepackage{mathtools}
\usepackage{amsthm}

\usepackage[capitalize,noabbrev]{cleveref}

\theoremstyle{plain}

\theoremstyle{definition}

\theoremstyle{remark}

\usepackage[textsize=tiny]{todonotes}

\usepackage{listings}
\definecolor{keywordcolor}{rgb}{0.7, 0.1, 0.1}   %
\definecolor{tacticcolor}{rgb}{0.0, 0.1, 0.6}    %
\definecolor{commentcolor}{rgb}{0.4, 0.4, 0.4}   %
\definecolor{symbolcolor}{rgb}{0.0, 0.1, 0.6}    %
\definecolor{sortcolor}{rgb}{0.1, 0.5, 0.1}      %
\definecolor{attributecolor}{rgb}{0.7, 0.1, 0.1} %

\input{lstjson}

\usepackage{caption}
\usepackage{subcaption}
\usepackage{graphicx}
\usepackage{float}
\usepackage{tikz}
\usetikzlibrary{arrows.meta, positioning, calc, shapes, backgrounds}
\usepackage[export]{adjustbox}

\usepackage{enumitem}

\usepackage{tcolorbox}

\icmltitlerunning{LeanTree: Accelerating White-Box Proof Search with Factorized States in Lean 4}

\begin{document}

\twocolumn[
\icmltitle{LeanTree: Accelerating White-Box Proof Search with\\Factorized States in Lean 4}

\icmlsetsymbol{equal}{*}

\begin{icmlauthorlist}
\icmlauthor{Matěj Kripner}{charles}
\icmlauthor{Michal Šustr}{ctu}
\icmlauthor{Milan Straka}{charles}
\end{icmlauthorlist}

\icmlaffiliation{charles}{Charles University, Faculty of Mathematics and Physics}
\icmlaffiliation{ctu}{Czech Technical University, Faculty of Electrical Engineering}

\icmlcorrespondingauthor{Matěj Kripner}{kripner@ufal.mff.cuni.cz}

\icmlkeywords{automated theorem proving, neural theorem proving, lean}

\vskip 0.3in
]

\printAffiliationsAndNotice{}  %

\begin{abstract}
Automated theorem proving (ATP) has been a classical problem in artificial intelligence since its inception, yet it remains challenging due to its vast state and action space. 
Large language models (LLMs) have recently emerged as a promising heuristic for ATP, but they lack correctness guarantees and thus require interaction with a proof verifier.
Such interactions typically follow one of two approaches: black-box interaction, which does not utilize intermediate proof states, or white-box approaches, which allow for incremental proof construction and examination of intermediate states. 
While black-box approaches have directly benefited from recent LLM advances, white-box methods have comparatively lagged behind.
In~this~paper, we address this gap by introducing \emph{LeanTree}, which consists of (i) a tool built in the Lean~4 language that factorizes complex proof states into simpler, independent branches, and (ii) a dataset of these factorized intermediate states. 
Our white-box tooling offers several advantages over black-box approaches: it simplifies evaluation, reduces necessary context, generates richer training data,
enables parallel search across multiple states, supports efficient reuse of states, and provides feedback in case of errors.
Our preliminary results hint that white-box approaches outperform black-box alternatives in some settings.
\end{abstract}

\section{Introduction}

Automated theorem proving (ATP) has long been a foundational task in artificial intelligence~\citep{gelernter1959}, with applications ranging from the formalization of mathematics~\citep{doorn2023sphere_eversion}, physics~\citep{tooby_smith2024formalization_physics} and chemistry~\citep{bobbin2024chemistry} to software verification~\citep{avigad2025cairo_zero}, cryptography~\citep{doussot2024cryptography}, SQL query verification~\citep{chu2018sql_queries}, and general reasoning~\citep{jiang2024leanreasoner}.

Despite extensive research using logical formalisms, ATP continues to face significant challenges due to the combinatorial explosion of its underlying state and action spaces~\citep{harrison2009handbook,baader2003description,barrett2018satisfiability} and the infinite branching factor introduced by exogenous term creation~\citep{trinh2024alpha_geometry}.

Recent advances in large language models (LLMs) capable of sophisticated reasoning in human languages~\citep{guo2025deepseek} have sparked interest in their application to ATP~\citep{ren2025deepseek}, serving as a powerful heuristic in navigating the proof search. 
Conversely, ATP and formal verification present an avenue for improving reliability and interpretability of LLMs~\citep{harmonic2023}.
Open-source tooling and datasets are crucial in leveraging this synergy.

The existing approaches for ATP can be categorized into two distinct methodologies. 
In \textit{white-box} generation, the prover interacts with a formal verifier iteratively, generating individual proof steps conditioned on the internal proof state, interleaving proof search with verification. 
In contrast, \textit{black-box} provers only use the formal verifier to verify the final proof, generating proof steps with no feedback about the validity of preceding steps or the current internal proof state.
Thus, in a~black-box search, it falls upon the proof step generator to deduce the evolving proof state, as this is a necessary precondition to determining the validity of a subsequent proof step.
This coupling of policy model and world model increases complexity of proof step generation, necessitating the use of more powerful models, while also requiring the proof generator to be retrained when modifying the formal verifier in case of upgrades or patches.

While black-box whole-proof generation methods directly benefit from recent advances in LLMs~\citep{wang2025kimina}, white-box tree-search methods have seen comparatively little improvement since their introduction.
Notably, the hypertree proof search approach introduced by \citet{lample2022hypertree} has not been replicated in open source or built upon, except for follow-up work by the same research group~\citep{gloeckle2024abel}.
We attribute this primarily to insufficient tooling and the lack of suitable datasets. 
Specifically, machine learning-based white-box techniques require nontrivial preprocessing of existing proofs to extract intermediate proof states. 
Additionally, the size of proof states in verification languages like Lean tends to increase as the proof search progresses, exacerbating the distribution shift for LLMs pretrained on general text corpora.
As such, it is desirable to simplify proof states by factorizing them into branches that can be solved independently.
Although \citet{lample2022hypertree} demonstrated the benefit of this white-box approach using factorized states, their tooling and dataset remain practically unusable for the ATP community (see Section~\ref{section:comparison}).

To address these challenges, we introduce \textit{LeanTree}, which is a) a tool for white-box interaction and data extraction in Lean 4~\citep{moura2021lean}, and b) a dataset of preprocessed proofs from Mathlib~\citep{mathlib2020} and DeepSeekProver-V1~\citep{xin2024deepseek} in a unified format.
Our contributions are as follows:
\begin{itemize}[noitemsep,topsep=2pt]
\item We build on top of Lean REPL~\citep{lean4repl} to enable programmatic proof search over factorized proof states, meaning that independent goals can be solved individually (Section~\ref{subsection:factorized_proof_states}).
To enable this, we detect dependence between goals caused by metavariable coupling (Section~\ref{subsection:metavariable_coupling}).
Other improvements over Lean REPL include better error reporting, richer information about open goals, and a new incremental strategy for proof verification in Lean REPL using the Lean's verification kernel (Appendix~\ref{appendix:incremental_proof_verification}).
\item We build a data extraction module by integrating Lean REPL, PaperProof~\citep{kovsharov2024paperproof}, and a custom algorithm for proof tree building and tactic simplification (Section~\ref{subsec:proof_tree_building}).
The extracted dataset also contains information about the current Lean context for each theorem and the distance to the end of the proof for each step.
\item We release the resulting suite of tools and datasets, dubbed \textit{LeanTree}, in a format directly accessible by the community\footnote{\url{https://github.com/Kripner/leantree}}\footnote{\url{https://huggingface.co/datasets/ufal/leantree}} with a user-friendly Python interface.
\item We present a preliminary experiment hinting that supplying information about the intermediate proof state to a pretrained model during proof search outperforms black-box generation.
\item We identify and prevent false-positive errors in proof search that result from the usage of library search tactics such as \texttt{apply?} (Appendix~\ref{appendix:library_search_tactics}).
For example, such errors are present in the proofs presented by DeepSeek-Prover-V2~\citep{ren2025deepseek}.
\end{itemize}

\section{Related work}

We give a brief overview of Lean, black-box and white-box ATP approaches, and existing datasets and benchmarks.

\textbf{Lean 4}~\citep{moura2021lean} is a leading formal language for theorem proving with applications both in the formalization of mathematics~\citep{doorn2023sphere_eversion} and in numerous other domains~\citep{tooby_smith2024formalization_physics,bobbin2024chemistry,doussot2024cryptography,jiang2024leanreasoner}.
One of the pleasant attributes of Lean is its extensibility, stemming from the fact that its theorem proving component is written in Lean itself, and the user can modify both its syntax and its semantics.
Such an approach is possible because Lean 4 is also a fully-fledged functional programming language.
This is in contrast to previous versions of Lean, where some of the existing Lean interaction tools had to bind directly to the Lean kernel written in C++.

\textbf{Black-box ATP.}
Automated theorem proving can be formulated as a text-to-text problem so that any existing approach based on a large language model (LLM) is directly applicable.
This approach to theorem proving was pioneered by Baldur~\citep{first2023baldur}.
Recently, a LLM with chain-of-thought refined using reinforcement learning was utilized by DeepSeek-Prover-V2~\citep{ren2025deepseek}, Kimina-Prover~\citep{wang2025kimina}, and InternLM-Math~\citep{ying2024internlm}, securing top places on relevant benchmarks.

\textbf{White-box ATP.}
In white-box ATP, the prover interacts with the verification system iteratively during a proof search, generating proof steps conditioned on the internal proof state, with consequences discussed in the introduction of this paper.
In neural theorem proving, this approach was pioneered by Holophrasm~\citep{whalen2016holophrasm} in Metamath~\citep{megill2019metamath}, using a UCT-based~\citep{kocsis2006uct} AND-OR tree search with a RNN sequence-to-sequence model for tactic enumeration and payoff prediction.
This approach was extended by GPT-f~\citep{polu2020gptf} and Evariste~\citep{lample2022hypertree}, with the latter additionally targeting Lean~3.
ABEL~\citep{gloeckle2024abel} extended Evariste with incremental improvements and Lean~4 compatibility, but did not release their code or dataset.

\textbf{Datasets.}
The standard library of Lean, Mathlib~\citep{mathlib2020}, offers more than 107k formal definitions and 220k formal proofs.
Together with numerous formalization projects created by the community, this offers a sizable human-written dataset for black-box theorem proving.

In contrast, white-box ATP approaches require data in a more preprocessed format that reveals intermediate proof states.
First, such a dataset was provided by LeanStep~\citep{han2021pact} in Lean 3 by extracting tactic invocation data from Mathlib.
LeanDojo~\citep{yang2023leandojo} extends this for Lean 4, additionally containing information about the premises used in each tactic application, suitable for training retrieval-augmented approaches.
Similar datasets are also offered by Pantograph~\citep{aniva2024pantograph} and lean-training-data~\citep{morrison2023lean-training-data}.
Additionally, Lean-Workbook~\citep{ying2024lean} and DeepSeek-Prover-V1~\citep{xin2024deepseek} provide large collections of automatically formalized proofs of various quality.
However, none of these datasets offers tactic proofs in a structured, simplified way, which we describe in this paper.

\textbf{Benchmarks.}
ATP benchmarks typically consist of a set of formalized theorem statements, optionally accompanied by their informal counterparts in natural language, where the objective is to generate a valid proof.
MiniF2F~\citep{zheng2021minif2f} offers 488 problems drawn from mathematical olympiad and undergraduate mathematics courses, formalized in parallel in Lean 3, Metamath, Isabelle, and HOL Light.
The benchmark was later ported to Lean 4 by \citet{yang2023leandojo}, with some formalization errors corrected by \citet{wu2024internlm2}.

Similarly, ProofNet~\citep{azerbayev2302proofnet} provides 371 problems formalized in Lean 3, together with their natural language statement and natural language proof, making it suitable also for evaluating autoformalization approaches. 
This benchmark was ported to Lean 4 by \citet{xin2024deepseek-v1.5}.

PutnamBench~\citep{tsoukalas2024putnambench} comprises of 1\,099 problems formalized in Lean 4, Isabelle, and Coq.
This benchmark is comparatively difficult, with DeepSeek-Prover-V2~\citep{ren2025deepseek}, the current state-of-the-art, attaining only 7.2 \% success rate with pass@1024.\footnote{\url{https://trishullab.github.io/PutnamBench/leaderboard.html} (retrieved Jun 2025)}

\textbf{Lean Tooling.}
The comparison of LeanTree with existing tools is given in Section~\ref{section:comparison}.

\section{Lean programmatic interaction}

We build on top of Lean REPL~\citep{lean4repl} to enable incremental proof execution.
In this section, we describe the LeanTree interaction module.

\textbf{Preliminaries.}~
In Lean, theorem proving consists of producing a term whose type is equal to the theorem type.
For example, consider the following theorem:
\begin{equation}
\begin{minipage}{0.9\linewidth}
\begin{lstlisting}[language=lean]
theorem sub_zero (a b : ℕ) (h : b = 0)
  : a - b = a
\end{lstlisting}
\end{minipage}
\label{lst:sub_zero}
\end{equation}
Proving Theorem~\ref{lst:sub_zero} consists of supplying a term of type \texttt{a - b = a} using available free variables \texttt{a}, \texttt{b} of type $\mathbb{N}$ and \texttt{h} of type \texttt{b = 0}.
While such a term can be written down explicitly, it is common to utilize Lean's \textit{tactic mode} to structure the proof in a format more aligned with human-written proofs.

Upon entering tactic mode using the \texttt{by} keyword, Lean creates a \textit{metavariable} with the desired type \texttt{a - b = a} and no value assigned yet.
Metavariables can be thought of as holes in a proof that have to be filled in before the proof is considered finished.
This gives us an intermediate state in the proof construction, where the proof term is type-correct but still contains holes to be filled in later.
The still unassigned metavariables are called \textit{open goals} so that each goal has a target type and a set of available free variables called \textit{hypotheses}.
Because goals are metavariables, the two terms are often used interchangeably.

We then proceed by applying a \textit{tactic} to the list of open goals.
A tactic is a procedure that assigns values to one or more open goals, with each of the values possibly containing new unassigned metavariables.
In this way, a tactic invocation transforms the list of open goals into a new list of open goals.
Typically, a tactic only affects the \textit{main goal} (defined to be the first goal), either solving it or reducing it to one or more simpler sub-goals.
The proof is finished once there are no more open goals, i.e., every metavariable has been assigned.

\subsection{LeanTree interaction module}

LeanTree offers a Python interface, where the user can initiate a proof search either from an individual theorem statement or from a Lean file.
After applying a Lean tactic to a proof state, LeanTree returns a~list of resulting sub-states.
This interface is suitable for tree search and backtracking as the interaction can be started from any intermediate proof state.
Additionally, given the high CPU requirements of Lean 4, LeanTree implements a dynamic pool of environments suitable for parallel execution.

\subsection{Factorized proof states}
\label{subsection:factorized_proof_states}

We follow \citet{lample2022hypertree} in factorizing proof states into independent goals, which can be explored and proven individually.
Note that in this section, we disregard metavariable coupling, which is then addressed in Section~\ref{subsection:metavariable_coupling}.

\begin{figure*}[t]
\centering
\includegraphics[width=0.7\linewidth]{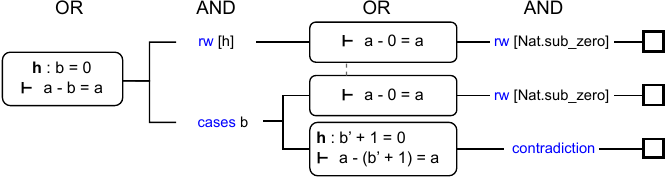}
\caption{
Example of an AND-OR proof tree for Theorem~\ref{lst:sub_zero}.
Nodes connected with a dashed line could potentially be merged and correspond to transpositions in board games.
This is possible thanks to factorization (see Section~\ref{subsection:factorized_proof_states}).
}
\label{fig:and-or}
\end{figure*}

The utility of proof state factorization has been identified by \citet{lample2022hypertree}, citing the ``re-use and parallelization in the proof search algorithm'' as their primary motivation.
An example of potential state reuse is shown in Figure~\ref{fig:and-or}, where the same node is reached using two different search paths.
To illustrate proof search parallelization, consider the following theorem:
\begin{equation}
\begin{minipage}{0.9\linewidth}
\begin{lstlisting}[language=lean]
theorem mul_eq_zero_iff (n m : ℕ)
  : n * m = 0 ↔ n = 0 ∨ m = 0
\end{lstlisting}
\end{minipage}
\label{lst:mul_eq_zero_iff}
\end{equation}
A possible strategy to prove Theorem~\ref{lst:mul_eq_zero_iff} is to branch out based on the values of $n$ and $m$.
We achieve this by executing the tactic \texttt{cases n <;> cases m}.
Since each of the two numbers can be either zero or a successor of another number following the Peano axioms, the resulting proof state consists of four goals.
For example, one of the states is $(n' + 1) \cdot 0 = 0 \leftrightarrow n' + 1 = 0 \lor 0 = 0$, corresponding to the case where $n$ is a successor and $m$ is zero.

To complete the proof, all four goals must be closed.
In a non-factorized setting, the proof state consisting of four goals is assessed by the policy and critic models as a whole.
Since tactics typically affect only the main goal, the last goal can only be worked on once all others have been closed.
In contrast, factorized states allow the search algorithm to focus on the goals individually, in any order the algorithm deems fit.

\looseness-1
As a final advantage, simplifying states reduces the difficulty of the policy task, deciding which tactic to apply next, and of the critic task, estimating the value of the current goal.
Note that proof states can grow arbitrarily complex during proof search since each state corresponds to a list of goals.

\textbf{AND-OR Search.}
Search over factorized states naturally leads to an AND-OR tree, where a prover selects a tactic in an OR node and subsequently has to prove all goals in an AND node. 
Figure~\ref{fig:and-or} shows an example where choosing either the \texttt{rw} tactic or the \texttt{cases} tactic in the root leads to a proof.
For a more in-depth introduction to proof search in Lean, we recommend \citep{aesop2023}.

\subsection{Metavariable coupling}
\label{subsection:metavariable_coupling}

Contrary to the simplified view presented in Section~\ref{subsection:factorized_proof_states}, it is not always possible to fully factorize the proof state into individual goals that can be proven independently.
Subsequently, nodes in the proof tree are allowed to contain more than one goal.
Specifically, this occurs when two goals share the same metavariable. 
For example, applying the transitivity lemma \texttt{Nat.le\_trans} to the goal $2 \leq 5$ produces goals $2 \leq \texttt{?m}$ and $\texttt{?m} \leq 5$ sharing a metavariable \texttt{?m} whose value has not yet been decided.\footnote{A third goal of type $\mathbb{N}$ is created for technical reasons.}
Subsequently, choosing a value for \texttt{?m} in one of the goals affects the provability of the other goal, and therefore the two goals cannot be explored independently.
For a more in-depth explanation of this phenomenon, we recommend~\citet{aniva2024pantograph}.

LeanTree detects all dependencies between goals caused by metavariables and offers strategies to deal with them during proof search.
By default, goals are factorized to the maximum extend allowed by metavariable coupling.

\subsection{Proof verification}

Originally, the Lean REPL did not utilize Lean verification kernel to type-check assignments introduced by tactic executions, leading to incorrect proofs being accepted in some cases.
Our prover managed to exploit these inconsistencies, finding 12 incorrect proofs on MiniF2F that passed verification in the REPL. We describe these incorrect proofs and recent approaches to fix Lean REPL in the appendix. Note that all proofs found by our prover are also independently verified directly using Lean, which guarantees their correctness.

\section{Supervised data extraction}

To enable machine learning-based approaches to imitate proofs without interfacing with Lean, LeanTree offers a data extraction module to create supervised data from existing Lean proofs.
The data extraction module is fully compatible with the interaction module in the sense that the extracted proof trees can be directly executed and verified.

\subsection{Proof tree building}
\label{subsec:proof_tree_building}

To achieve compatibility with the interaction module, the proof states extracted from supervised data must be factorized into independent branches as described in Section~\ref{subsection:factorized_proof_states}.
This process yields a proof tree where each node contains a list of goals (typically a singleton list), and each edge corresponds to a tactic application.

\begin{figure*}[t]
    \centering
    \includegraphics[width=.9\hsize]{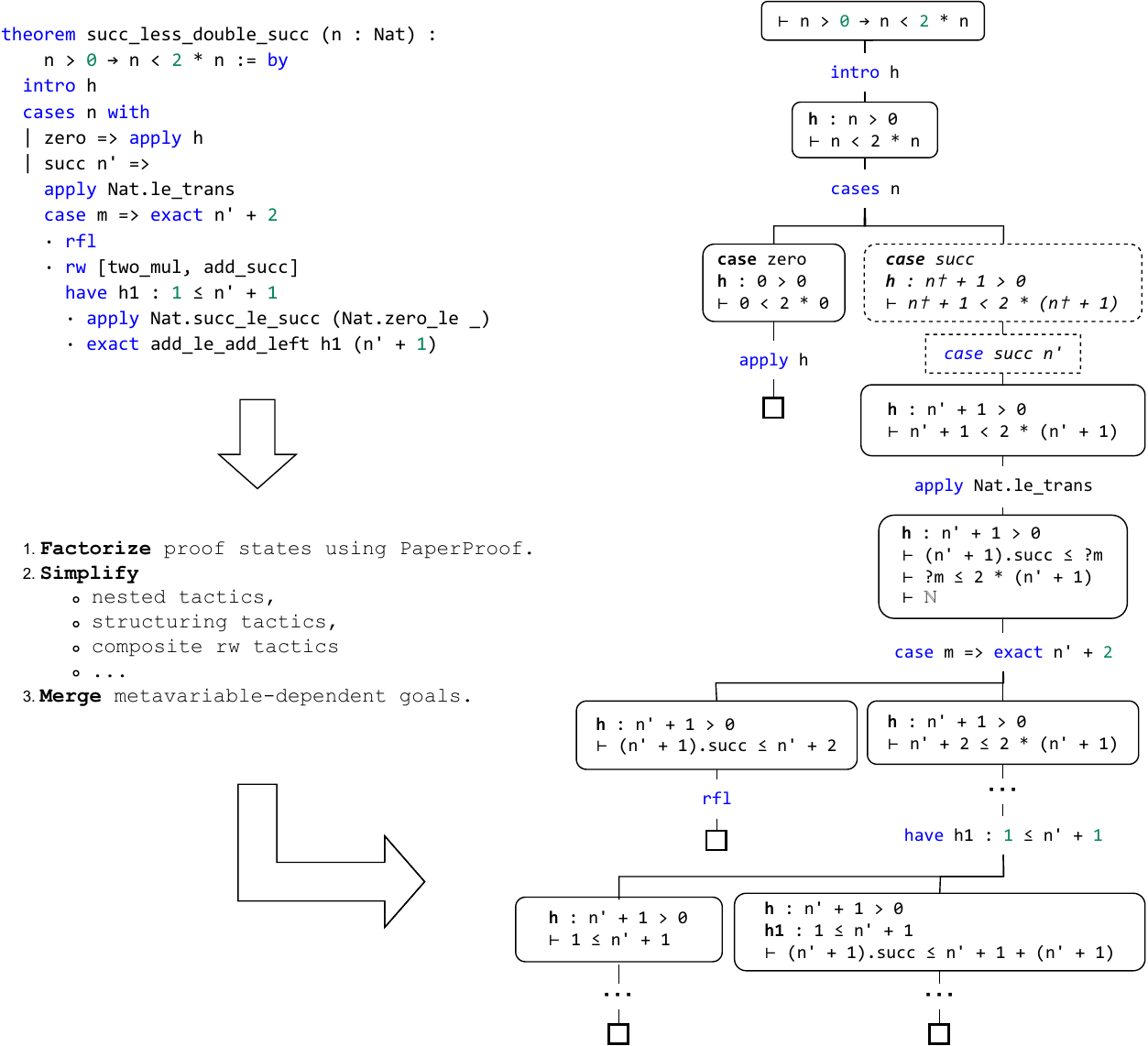}
    \caption{Proof tree builder in LeanTrees transforms a Lean proof with complex structure (left) into a proof tree (right) with simple states (lists of goals that cannot be split further) and transitions (tactics). Synthetic elements in the proof tree are marked using dashed border.}
    \label{fig:succ_less_double_succ}
\end{figure*}

To illustrate some challenges in building a proof tree, consider the Lean proof and its corresponding proof tree shown in Figure~\ref{fig:succ_less_double_succ}, where we prove $n < 2n$ under the condition $n > 0$.
For the purposes of this illustration, we do not utilize powerful tactics like \texttt{linarith}, which would solve this trivial theorem nearly automatically.
Instead, we illustrate branching, metavariable coupling, and the introduction of a new hypothesis.

Note that in the original Lean proof in Figure~\ref{fig:succ_less_double_succ}, tactics are nontrivially structured via branching using \texttt{cases}, focusing a goal via the center-dot operator, and merging multiple \texttt{rw} commands into one.
Additionally, each Lean tactic consumes the set of currently open goals -- the unfactorized proof state -- and produces a new set of open goals -- a new proof state.
In contrast, each edge of the constructed proof tree corresponds to a simple tactic applied to a minimal list of goals that cannot be split further due to metavariable coupling.

To illustrate metavariable coupling, observe that applying \texttt{Nat.le\_trans} yields three goals that are not split due to a shared metavariable \texttt{?m}.
However, once the metavariable is assigned using \texttt{exact}, the remaining two goals are now independent and can be split.

Finally, note that in the proof tree, the tactic \texttt{case succ n'} and its parent node are synthetic in the sense that they do not have a counterpart in the original proof.
This synthetic tactic is necessary to assign the name \texttt{n'} to the otherwise inaccessible variable -- denoted \texttt{n\textdagger} before the assignment -- so that it can be referenced by later tactics.

Other Lean features not shown in Figure~\ref{fig:succ_less_double_succ} that further complicate proof tree construction include tactic combinators such as \texttt{try}, \texttt{iterate}, and \texttt{any\_goals}, where the set of affected goals is determined in runtime, constructs such as \texttt{all\_goals} and \texttt{<;>} that go against the tree formulation by operating on multiple goals, or tactics like \texttt{switch} and \texttt{rotate\_left} that change the order of open goals.

We implement the transformation illustrated in Figure~\ref{fig:succ_less_double_succ} in three stages described in the following three subsections.

\subsubsection{Singleton trees}
In the first stage of proof tree building, a Lean proof is transformed into a \textit{singleton tree}, where each node contains a single goal, by matching tactic applications with the goals they affect.
This is nontrivial, because, in general, a tactic application transforms a list of open goals into an entirely new list of empty goals without any guarantees about the transformation.

To build the singleton tree, we integrate Lean REPL with the BetterParser module of PaperProof~\citep{kovsharov2024paperproof}, which solves the analogous problem for the purposes of visualization.

\subsubsection{Tactic simplification}
\label{subsubsec:tactic_simplification}
In the second stage, the singleton tree is modified so that complex tactics are broken down into simpler ones and each tactic is a valid self-contained proof step.
Here we describe some of these modifications.

\textbf{Nested Tactics.}
Tactics parametrized by a term can themselves contain nested tactic blocks.
For example, consider the following tactic from the Complex Analysis module in Mathlib.
\begin{equation}
    \begin{minipage}{0.9\linewidth}
        \begin{lstlisting}[language=lean]
exact ⟨by 
 rw [aeval_algHom_apply, hw, map_zero],
 rfl⟩
\end{lstlisting}
    \end{minipage}
    \label{lst:exact_with_by_block}
\end{equation}
In Listing~\ref{lst:exact_with_by_block}, the \texttt{exact} tactic is parametrized by a term which contains a tactic block introduced using the \texttt{by} keyword.
We break down this complex proof step by masking the nested tactic block and instead spawning a child goal, yielding the following series of simpler proof steps.
\begin{equation}
    \begin{minipage}{0.9\linewidth}
        \begin{lstlisting}[language=lean]
exact ⟨by sorry, rfl⟩
rw [aeval_algHom_apply]
rw [hw]
rw [map_zero]
\end{lstlisting}
    \end{minipage}
    \label{lst:exact_broken_down}
\end{equation}
Note that the \texttt{rfl} tactic is not masked since it is used as a term instead of as a tactic.

\textbf{Structuring tactics.}
The \texttt{cases} and \texttt{induction} tactics introduce syntactic branching by partitioning a proof based on the value of an inductive type.
The respective tactic optionally contains a solving branch for each constructor of such type.
For example, the \texttt{cases} tactic in Listing~\ref{fig:succ_less_double_succ} provides separate proofs for the \texttt{zero} and \texttt{succ} constructors of variable \texttt{n}.
This \texttt{cases} tactic spans 10 lines, which is contrary to our effort of breaking down the proof into simple steps.
For this reason, we move the individual proof branches from the \texttt{cases} tactic into their separate proof steps.

However, branches corresponding to parametrized constructors can optionally assign names to some of the constructors' parameters.
For example, in Listing~\ref{fig:succ_less_double_succ}, the \texttt{succ} constructor is parametrized by a~single natural number which is given the name \texttt{n'}.
When the \texttt{cases} tactic is broken down, the name \texttt{n'} has to be assigned manually, since Lean disallows referring to variables with no explicitly assigned name.
We achieve this by adding a synthetic tactic \texttt{case} which has no direct tactic counterpart in the original proof.
Note that the same effect could be achieved using the \texttt{rename} tactic, which is, however, rare in human-written proofs.

\textbf{Merged \texttt{rw} tactics.}
In Lean, subsequent \texttt{rw} tactics can be merged into a single tactic, which is commonplace in human-writen proofs.
We partition such composite tactics back into individual \texttt{rw} tactics.
This is also the case for the \texttt{rwa} tactic, where we additionally add the \texttt{assumption} tactic to preserve semantics.

\subsubsection{Merging metavariable-dependent goals}
To obtain the final proof tree, sibling nodes in a singleton tree are merged if they share a metavariable.
This corresponds to the fact that metavariable-dependent goals cannot be explored independently.

During this merging process, the entire proof tree is executed in Lean REPL starting from the root state and following each tree edge, verifying that each proof branch ends in a proven state.
In this way, we ensure the correctness of the whole proof tree building process.
Proof trees that fail this verification are not included in the final dataset (cf. Section~\ref{subsec:leantree_dataset}).

\subsection{LeanTree dataset}
\label{subsec:leantree_dataset}
We release the extracted proof tree data in an unified format from two sources: 1) a recent version of Mathlib 4,\footnote{\texttt{leanprover/lean4:v4.19.0}, Apache-2.0 license} the standard library of human-written proofs in Lean, and 2) a collection of 27.5K proofs autoformalized by DeepSeek-Prover-V1.\footnote{\url{https://huggingface.co/datasets/deepseek-ai/DeepSeek-Prover-V1}, MIT License}
Importantly, each sample in the LeanTree dataset corresponds to a Lean file rather than just an individual theorem.
This is necessary to capture the structure of a real-world Lean project like Mathlib where a proof can depend on any definition located above it in the source file.

Each file in the LeanTree dataset contains a list of theorems, and each theorem contains a list of all tactic proofs in its proof term.
Note that there can be more than one tactic proof for a theorem if its proof contains more than one non-nested \texttt{by}-blocks.
For each tactic proof, LeanTree then contains a proof tree with nodes corresponding to factorized proof states and edges corresponding to tactic applications.

To demonstrate a possible use case for proof trees, the dataset also contains the size and depth for each proof tree node.
These can serve as objectives for a critic model in various proof search algorithm.

Additionally, the LeanTree dataset contains information about the surrounding context, namely the list of imported modules for each Lean file and the list of open namespaces for each theorem.
The correspondence between samples in the dataset and the underlying Lean repository is given by character offsets specifying the span of each theorem, proof, and tactic execution.

Overall, LeanTree contains 74\,706 factorized tactic proofs from Mathlib and 26\,201 from DeepSeek-Prover-V1. Since Lean was not designed to enable factorized proof tree search out-of-the-box, there are a large number of small technical challenges to overcome during the proof tree building.
While we are continually working on perfecting this process, not all tactic proofs can currently be converted.
Specifically, 23.0\% of tactic proofs in Mathlib and 4.7\% in DeepSeek-Prover-V1 were not converted.
We note that for Mathlib, 28.6\% of the issues stem from the usage of \texttt{calc} and \texttt{conv} tactics that fundamentally change the structure of a proof.

\section{Comparison to existing tools} \label{section:comparison}
Several tools have recently emerged both for programmatic interaction with Lean and machine learning data extraction from human-written Lean proofs.
In this section, we compare them to LeanTree.
We argue that LeanTree is unique in its data extraction capabilities.  %

\textbf{Evariste~\citep{lample2022hypertree}} corresponds most closely to LeanTree. Evariste enables factorizing proof state into individual goals, which can be explored independently.
In addition, Evariste detects metavariable coupling and does not split metavariable-dependent goals, similar to LeanTree.

However, the Evariste dataset has not been released and the tool itself is available only in a limited, nonexecutable format.\footnote{Specifically, the codebase states: ``code does NOT run out of the box as we removed references to Meta-internal systems''~\url{https://github.com/facebookresearch/Evariste}}
This makes it challenging to build on top of Evariste's contributions.

Importantly, Evariste was only implemented for Lean 3, which is now deprecated.
Since its core logic is implemented using C++, directly binding to Lean~3 internals, its migration to Lean 4 would be difficult.
In contrast, the core logic of LeanTree is implemented in Lean 4 itself, which is a common Lean 4 paradigm~\citep{moura2021lean}.
Furthermore, LeanTree builds on top of the semi-official Lean REPL project, and it is therefore set to benefit from future improvements by the community.

\textbf{LeanDojo~\citep{yang2023leandojo}} 
offers both an interaction module and a supervised data extraction module.
The extracted data also contains information about the premises used in each tactic application, enabling training of retrieval-augmented generation models.
The extracted datasets are available for both Lean 3 and Lean 4.

However, LeanDojo does not offer proof state factorization, tactic simplification, or proof tree building.
First, this means that a tactic can only be applied to a whole proof state consisting of a list of goals, yielding a new set of goals.
This prevents parallel search over independent goals.

Second, tactics in the LeanDojo dataset are left in the form in which they appear in the abstract syntax tree of a Lean source code.
This includes complex or nested tactics that span multiple lines.
For example, the \texttt{cases} tactic shown in Figure~\ref{fig:succ_less_double_succ} is left unaltered.
The user of the dataset can then choose to filter out such samples, losing valuable training data, or keep them in the training set, but increase the difficulty of the modeling task and reduce the granularity of the inference-time proof search.

Third, LeanDojo does not contain information about the sub-proof size and sub-proof depth for each node, and provides tactics in an unstructured format that is not conducive to proof tree building.

\textbf{Pantograph~\citep{aniva2024pantograph}}
In contrast to LeanDojo, Pantograph's interaction and data extraction modules offer proof state factorization, meaning that independent goals can be explored independently.
Pantograph also provides information about which goals are coupled via a shared metavariable.

\begin{figure*}[t]
  \centering
  \begin{minipage}[c]{.45\linewidth}
    \centering
    \begin{tabular}{ll}
      \toprule
      Approach & MiniF2F-test \\
      \midrule
      Whole-proof     & \ \ 9.59 \% $\pm$ 0.71   \\
      Black-box rollout      & \ \ 5.32 \% $\pm$ 0.37   \\
      White-box rollout     & 18.36 \% $\pm$ 0.60   \\
      \midrule
      \textit{Best-first search}$^{*}$   & \textit{26.23 \%} \\
      \bottomrule
    \end{tabular}
    \smallskip
    
    \noindent{\footnotesize $^{*}$Reported by \citet{azerbayev2023llemma}.}
    \label{fig:minif2f_table}
  \end{minipage}%
  \hfill
  \begin{minipage}{.49\linewidth}
    \includegraphics[width=\textwidth,valign=M]{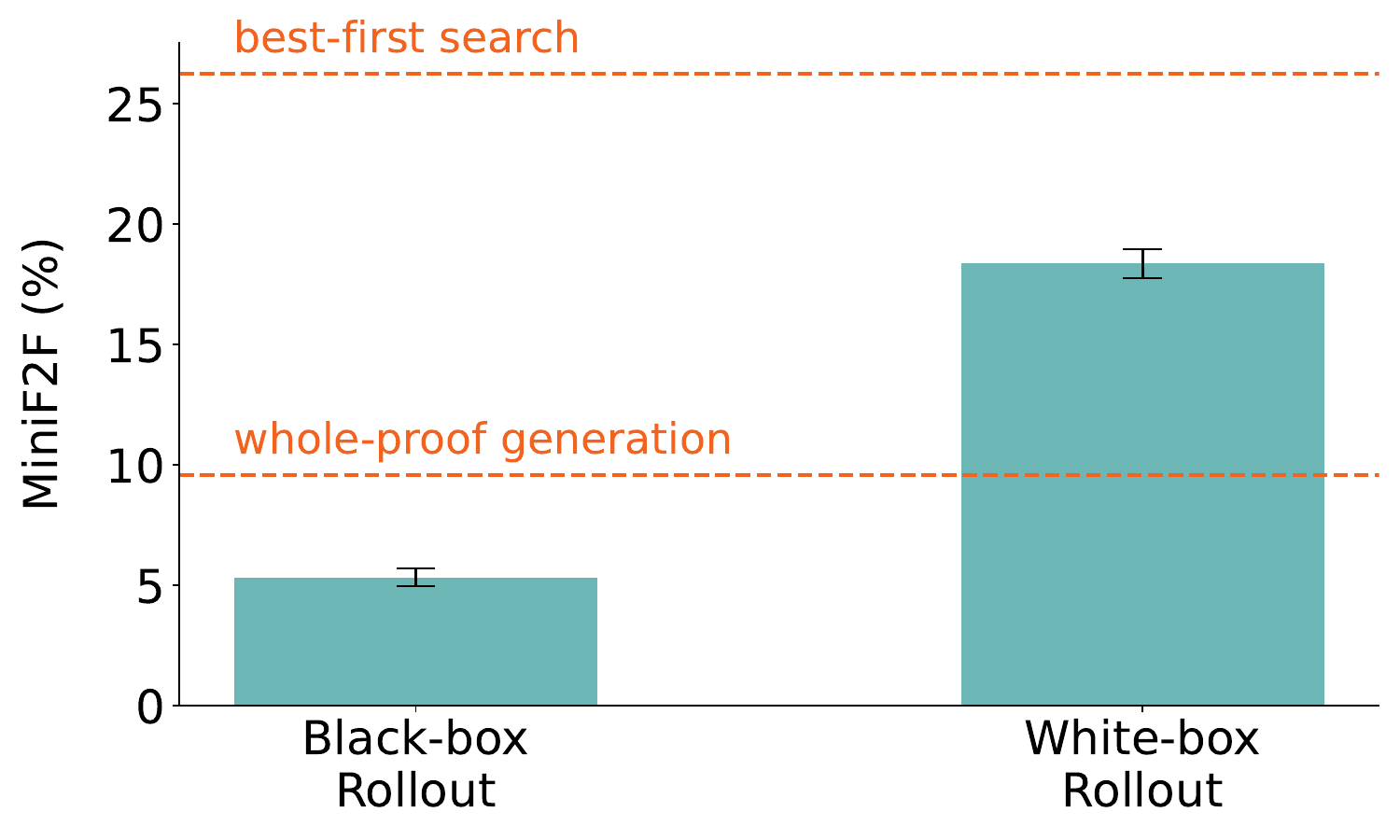}
    \label{fig:minif2f_plot}
  \end{minipage}
  \hfill~
  \caption{Performance on MiniF2F using linear rollouts with Llemma-7B. 
  Error bars show standard deviation over 5 runs.
  }
  \label{fig:minif2f_table_and_plot}
\end{figure*}

However, tactic simplification and proof tree building are not supported by the data extraction module, leading to the same drawbacks as described for the case of LeanDojo.
Information about sub-proof size and sub-proof depth is also not provided.

In addition, the Pantograph API is not well suited for gradually building the current Lean context by interleaving definitions and theorems that can depend on any previously declared ones.
Instead, theorems are treated as independent units.

\section{Experiments}

We investigate the utility of white-box approach in proof search using the Llemma-7B model~\citep{azerbayev2023llemma} (MIT license), reporting the success rate on MiniF2F test set~\citep{zheng2021minif2f} modified for Lean 4 by \citet{wu2024internlm2}.
For each input theorem, we run $N = 10$ independent linear rollouts, starting from an empty proof and sequentially sampling tactics proposed by the model for a maximum of $M = 25$ steps.
The search is successful if the concatenated tactics from a rollout are a valid proof of the input theorem.
If no proof is found during the $N$ rollouts, the search is unsuccessful.

In Figure~\ref{fig:minif2f_table_and_plot}, we report the results for the black-box and white-box rollouts averaged over 5 runs with error bars of one standard deviation.
We additionally report the success rate of black-box whole proof generation, where the model is prompted to generate the entire proof at once.
Lastly, we include the success rate of proof generation using best-first search~\citep{polu2020gptf} reported by \citet{azerbayev2023llemma}, which performs white-box rollouts in a tree fashion and is therefore a natural upper bound on the white-box rollout performance.

Our results show that in our specific setting, supplying the prover with internal proof states improves proof search performance.
Interestingly, black-box linear rollouts achieve worse performance than black-box whole-proof generation, indicating that constraining the model to generate only one tactic at a time hurts its performance.
However, rollouts with access to the internal state consistently outperform both black-box methods.

We run all our experiments on a single node equipped with 8 AMD MI210 accelerators and 192 CPU nodes, parallelizing both model inference and Lean interaction.
In total, our experiments necessitated 59 GPU-hours and 450 CPU-hours.

\section{Conclusion}
In this work, we introduced \textit{LeanTree}, a novel white-box tool designed to accelerate white-box approaches to automated theorem proving in Lean~4 via factorized proof states and structured proof trees.
Our approach addresses limitations of existing black-box methods by leveraging intermediate proof states, enabling parallelized and efficient proof search, and providing structured feedback during theorem proving.
LeanTree's integration with Lean REPL ensures that it will continue to benefit from patches and fixes by the Lean community.

Additionally, we released a unified dataset of factorized intermediate proof states derived from Mathlib and DeepSeek-Prover-V1.
A limitation of LeanTree is that currently not all tactic proofs can be converted. 

Finally, our preliminary experiments hint that white-box approaches, facilitated by LeanTree, can outperform black-box methods in automated theorem proving.
We hope that LeanTree serves as a stepping stone toward more efficient, parallelizable, and reliable proof search and encourages further research into white-box automated theorem proving.

\section*{Acknowledgements}
Matěj Kripner was supported by the grant no. 25-18031S of the Czech Science Foundation (GAČR).
This research was partially supported by SVV project number 260 821.
We acknowledge VSB – Technical University of Ostrava, IT4Innovations National Supercomputing Center, Czech Republic, for awarding this project access to the LUMI supercomputer, owned by the EuroHPC Joint Undertaking, hosted by CSC (Finland) and the LUMI consortium through the Ministry of Education, Youth and Sports of the Czech Republic through the e-INFRA CZ (grant ID: 90254).
The authors thank Martin Suda, Matej Straka, Tomáš Čížek, Radovan Haluška, Martin Schmid, and the Lean community for helpful suggestions and corrections.

\bibliography{main}
\bibliographystyle{icml2025}

\newpage
\appendix
\onecolumn

\section{False-positives with library search tactics}
\label{appendix:library_search_tactics}
The \texttt{apply?} tactic\footnote{\url{https://lean-lang.org/doc/reference/latest/Tactic-Proofs/Tactic-Reference/\#apply\_\_\_}} is intended for searching the Lean library for lemmas applicable in the current context during interactive proof construction, similar to \texttt{exact?} and \texttt{rw?}.
However, the \texttt{apply?} tactic must not be left in a final proof, as it has the same semantics as the \texttt{sorry} keyword, leaving such a proof incomplete.

For example, the proofs of \texttt{putnam\_2005\_a4} and \texttt{putnam\_2007\_b4} presented by DeepSeek-Prover-V2~\citep{ren2025deepseek} utilize the \texttt{apply?} tactic and are thus incorrect.
This highlights the potential of reinforcement learning to exploit unchecked ambiguities in a proving environment.
It also hints as to why a smaller DeepSeek-Prover-V2-7B model beat DeepSeek-Prover-V2-671B in some contexts related to cardinal numbers, as the former might have learned to exploit the \texttt{apply?} tactic while the latter did not (cf. Section \textit{Skill Discovery by Reinforcement Learning} in \citet{ren2025deepseek}).

In LeanTree, we prevent such errors by banning the usage of \texttt{apply?} by default, leaving it as opt-in for users who explicitly want to integrate it into their proving system.

\section{Incremental proof verification}
\label{appendix:incremental_proof_verification}

Originally, the Lean REPL did not utilize the Lean verification kernel to type-check assignments introduced by tactic applications, leading to incorrect proofs being accepted in some cases.
Our prover managed to exploit these inconsistencies, finding 12 incorrect proofs in the MiniF2F validation set that passed verification in the REPL when using the tactic mode.
One of these incorrect proofs is listed below.

\begin{minipage}{\textwidth}
\begin{lstlisting}[language=lean]
theorem mathd_algebra_422 (x : ℝ) (σ : Equiv ℝ ℝ) (h₀ : ∀ x, σ.1 x = 5 * x - 12)
    (h₁ : σ.1 (x + 1) = σ.2 x) : x = 47 / 24 := by
  rw [h₀] at h₁
  by_contra! h
  apply h.symm
  simpa using h _
\end{lstlisting}
\end{minipage}

Lean correctly rejects the above proof with the error: ``\textit{don't know how to synthesize placeholder}''.
However, when executing the proof in the tactic mode using the old version of Lean REPL, the proof passes as correct.

In a recent update,\footnote{Published on GitHub on Apr 10, 2025.} Lean REPL was extended to type-check the full proof term after each tactic application, mitigating all false positives in the verification process, at the cost of introducing some drawbacks detailed below.
We propose a different strategy: Instead of verifying the type-correctness of the whole proof term, type-check only the new assignments introduced by a tactic application.

To illustrate this difference, assume that we are searching for a proof of type $T$ and have already constructed a term $t : T$ which still contains unassigned metavariables of types $G_1, \ldots, G_n$.
Next, assume that we apply a tactic assigning terms $g_1 : G_1, \ldots, g_n : G_n$ to the metavariables in $t$.
The current strategy of Lean REPL is to type-check the whole assignment $t : T$, which contains the new terms $g_1, \ldots, g_n$ as subterms.

This leads to issues when the proof bifurcates into two branches.
For example, consider applying the tactic \texttt{have k := sorry} in the REPL's tactic mode.\footnote{This is the standard way to use the \texttt{have} tactic in the REPL -- see for example \url{https://github.com/leanprover-community/repl/blob/master/test/have_by_sorry.in}}
This corresponds to the assignment of the term \texttt{((fun k $\mapsto$ ?\_) sorry)} to the metavariable corresponding to the main goal, where \texttt{?\_} is a new unassigned metavariable.
Lean REPL then enables solving the \texttt{?\_} and \texttt{sorry} parts independently.
However, assignments made in one of these two branches are not propagated to the other one, meaning that, from the perspective of each branch, the root proof term will always contain at least one hole introduced by an unassigned metavariable or the \texttt{sorry} keyword.
This leads to the kernel always failing during a type check for the duration of this proof.
The end effect is that the current verification strategy leads to false negatives when the proof being verified contains at least one syntactic branching.

In contrast, the proposed strategy implemented in LeanTree type-checks only the new assignments $g_1 : G_1, \ldots, g_n : G_n$ introduced by a tactic application.
The type correctness of the whole proof term $t : T$ then follows transitively.
Most importantly, this new verification strategy mitigates the false negatives described in the previous paragraph.
Additionally, the new strategy also decreases the time complexity of verifying a proof from quadratic to linear since each sub-assignment is only verified once.

Note that all proofs found by our prover are independently verified directly using Lean, bypassing Lean REPL together with the new verification algorithm described in this section.

\section{Experiments setup}
In all experiments, we prompt the Llema-7B model using the following prompt.

\begin{tcolorbox}[colback=white, colframe=black, sharp corners, boxrule=0.5pt]
\texttt{%
Complete the given Lean 4 code.\\
Directly output the completed Lean 4 code without any additional text, comments, or reasoning.
}
\end{tcolorbox}

When performing linear rollouts, we manually select only the first tactic from the model output.

During generation, we disable sampling of the \texttt{sorry} and \texttt{admit} keywords, together with syntax for comment sections.

\clearpage
\section{Dataset schema}
The dataset is distributed in two files corresponding to the two data sources:
\begin{itemize}
\item \texttt{lean-trees\_mathlib.jsonl}
\item \texttt{lean-trees\_deepseek-prover-v1.jsonl}
\end{itemize}
Both of these files share the following unified schema.

\paragraph{}

\begin{minipage}{\textwidth}
\begin{lstlisting}[language=json]
<sample> ::= {
  "path": <string>,
  "imports": [<string>],
  "theorems": [<error> | {
    "span": <span>,
    "name": <string?>,
    "context": [<string>],
    "by_blocks": [{
      "tree": <error> | {
        "root": <proof_node>
      }
    }]
  }]
}

<proof_node> ::= {
  "id": <string>,
  "proof_size": <int>,
  "proof_depth": <int>,
  "tactic": {
    "tactic_string": <string>,
    "span": <span>,
    "children": [<string>],
    "tactic_depends_on": [<string>]
  }
  "state": {
    "goals": [{
      "tag": <string?>,
      "type": <string>,
      "hypotheses": [{
        "type": <string>,
        "user_name": <string>,
        "value": <string?>
      }]
    }]
  }
}
    
<span> ::= {
  "start": <int>,
  "finish": <int>
}

<error> ::= {
  "error": <string>
}
\end{lstlisting}
\end{minipage}

\end{document}

%% file: lstjson.tex
\colorlet{punct}{red!60!black}
\definecolor{background}{HTML}{FFFFFF}
\definecolor{delim}{RGB}{20,105,176}
\colorlet{numb}{magenta!60!black}

\lstdefinelanguage{json}{
    basicstyle=\small\ttfamily,
    numbers=left,
    numberstyle=\scriptsize,
    stepnumber=1,
    numbersep=8pt,
    showstringspaces=false,
    breaklines=true,
    frame=lines,
    backgroundcolor=\color{background},
    literate=
     *{0}{{{\color{numb}0}}}{1}
      {1}{{{\color{numb}1}}}{1}
      {2}{{{\color{numb}2}}}{1}
      {3}{{{\color{numb}3}}}{1}
      {4}{{{\color{numb}4}}}{1}
      {5}{{{\color{numb}5}}}{1}
      {6}{{{\color{numb}6}}}{1}
      {7}{{{\color{numb}7}}}{1}
      {8}{{{\color{numb}8}}}{1}
      {9}{{{\color{numb}9}}}{1}
      {:}{{{\color{punct}{:}}}}{1}
      {,}{{{\color{punct}{,}}}}{1}
      {\{}{{{\color{delim}{\{}}}}{1}
      {\}}{{{\color{delim}{\}}}}}{1}
      {[}{{{\color{delim}{[}}}}{1}
      {]}{{{\color{delim}{]}}}}{1},
}

%% file: main.bbl
\begin{thebibliography}{41}
\providecommand{\natexlab}[1]{#1}
\providecommand{\url}[1]{\texttt{#1}}
\expandafter\ifx\csname urlstyle\endcsname\relax
  \providecommand{\doi}[1]{doi: #1}\else
  \providecommand{\doi}{doi: \begingroup \urlstyle{rm}\Url}\fi

\bibitem[Achim \& Tenev(2023)Achim and Tenev]{harmonic2023}
Achim, T. and Tenev, V.
\newblock Harmonic: Building mathematical superintelligence.
\newblock \url{https://harmonic.fun/}, 2023.
\newblock Accessed: 2025-05-14.

\bibitem[Aniva et~al.(2024)Aniva, Sun, Miranda, Barrett, and
  Koyejo]{aniva2024pantograph}
Aniva, L., Sun, C., Miranda, B., Barrett, C., and Koyejo, S.
\newblock Pantograph: A machine-to-machine interaction interface for advanced
  theorem proving, high level reasoning, and data extraction in lean 4.
\newblock \emph{arXiv preprint arXiv:2410.16429}, 2024.

\bibitem[Avigad et~al.(2025)Avigad, Goldberg, Levit, Seginer, and
  Titelman]{avigad2025cairo_zero}
Avigad, J., Goldberg, L., Levit, D., Seginer, Y., and Titelman, A.
\newblock A proof-producing compiler for blockchain applications.
\newblock \emph{J. Autom. Reason.}, 69\penalty0 (2), April 2025.
\newblock ISSN 0168-7433.
\newblock \doi{10.1007/s10817-025-09723-y}.
\newblock URL \url{https://doi.org/10.1007/s10817-025-09723-y}.

\bibitem[Azerbayev et~al.(2023{\natexlab{a}})Azerbayev, Piotrowski, Schoelkopf,
  Ayers, Radev, and Avigad]{azerbayev2302proofnet}
Azerbayev, Z., Piotrowski, B., Schoelkopf, H., Ayers, E.~W., Radev, D., and
  Avigad, J.
\newblock Proof{N}et: Autoformalizing and formally proving undergraduate-level
  mathematics (2023).
\newblock \emph{URL https://arxiv. org/abs/2302.12433}, 2023{\natexlab{a}}.

\bibitem[Azerbayev et~al.(2023{\natexlab{b}})Azerbayev, Schoelkopf, Paster,
  Santos, McAleer, Jiang, Deng, Biderman, and Welleck]{azerbayev2023llemma}
Azerbayev, Z., Schoelkopf, H., Paster, K., Santos, M.~D., McAleer, S., Jiang,
  A.~Q., Deng, J., Biderman, S., and Welleck, S.
\newblock Llemma: An open language model for mathematics.
\newblock \emph{arXiv preprint arXiv:2310.10631}, 2023{\natexlab{b}}.

\bibitem[Baader(2003)]{baader2003description}
Baader, F.
\newblock \emph{The description logic handbook: Theory, implementation and
  applications}.
\newblock Cambridge university press, 2003.

\bibitem[Barrett \& Tinelli(2018)Barrett and
  Tinelli]{barrett2018satisfiability}
Barrett, C. and Tinelli, C.
\newblock Satisfiability modulo theories.
\newblock \emph{Handbook of model checking}, pp.\  305--343, 2018.

\bibitem[Bobbin et~al.(2024)Bobbin, Sharlin, Feyzishendi, Dang, Wraback, and
  Josephson]{bobbin2024chemistry}
Bobbin, M.~P., Sharlin, S., Feyzishendi, P., Dang, A.~H., Wraback, C.~M., and
  Josephson, T.~R.
\newblock Formalizing chemical physics using the lean theorem prover.
\newblock \emph{Digital Discovery}, 3\penalty0 (2):\penalty0 264--280, 2024.

\bibitem[Chu et~al.(2018)Chu, Murphy, Roesch, Cheung, and
  Suciu]{chu2018sql_queries}
Chu, S., Murphy, B., Roesch, J., Cheung, A., and Suciu, D.
\newblock Axiomatic foundations and algorithms for deciding semantic
  equivalences of sql queries.
\newblock \emph{Proc. VLDB Endow.}, 11\penalty0 (11):\penalty0 1482–1495,
  July 2018.
\newblock ISSN 2150-8097.
\newblock \doi{10.14778/3236187.3236200}.
\newblock URL \url{https://doi.org/10.14778/3236187.3236200}.

\bibitem[Community(2025)]{lean4repl}
Community, L.
\newblock Lean 4 repl.
\newblock \url{https://github.com/leanprover-community/repl}, 2025.
\newblock Accessed: 2025-04-28.

\bibitem[Community(2020)]{mathlib2020}
Community, M.
\newblock The lean mathematical library.
\newblock In \emph{Proceedings of the 9th ACM SIGPLAN International Conference
  on Certified Programs and Proofs}, CPP 2020, pp.\  367–381, New York, NY,
  USA, 2020. Association for Computing Machinery.
\newblock ISBN 9781450370974.
\newblock \doi{10.1145/3372885.3373824}.
\newblock URL \url{https://doi.org/10.1145/3372885.3373824}.

\bibitem[Doussot(2024)]{doussot2024cryptography}
Doussot, G.
\newblock Cryptography experiments in lean 4: {SHA}-3 implementation.
\newblock Cryptology {ePrint} Archive, Paper 2024/1880, 2024.
\newblock URL \url{https://eprint.iacr.org/2024/1880}.

\bibitem[First et~al.(2023)First, Rabe, Ringer, and Brun]{first2023baldur}
First, E., Rabe, M.~N., Ringer, T., and Brun, Y.
\newblock Baldur: Whole-proof generation and repair with large language models.
\newblock In \emph{Proceedings of the 31st ACM Joint European Software
  Engineering Conference and Symposium on the Foundations of Software
  Engineering}, pp.\  1229--1241, 2023.

\bibitem[Gelernter(1959)]{gelernter1959}
Gelernter, H.~L.
\newblock Realization of a geometry theorem-proving machine.
\newblock In \emph{Proceedings of the First International Conference on
  Information Processing (IFIP)}, pp.\  273--281, 1959.

\bibitem[Gloeckle et~al.(2024)Gloeckle, Limperg, Synnaeve, and
  Hayat]{gloeckle2024abel}
Gloeckle, F., Limperg, J., Synnaeve, G., and Hayat, A.
\newblock {ABEL}: Sample efficient online reinforcement learning for neural
  theorem proving.
\newblock In \emph{The 4th Workshop on Mathematical Reasoning and AI at
  NeurIPS'24}, 2024.

\bibitem[Guo et~al.(2025)Guo, Yang, Zhang, Song, Zhang, Xu, Zhu, Ma, Wang, Bi,
  et~al.]{guo2025deepseek}
Guo, D., Yang, D., Zhang, H., Song, J., Zhang, R., Xu, R., Zhu, Q., Ma, S.,
  Wang, P., Bi, X., et~al.
\newblock Deepseek-r1: Incentivizing reasoning capability in llms via
  reinforcement learning.
\newblock \emph{arXiv preprint arXiv:2501.12948}, 2025.

\bibitem[Han et~al.(2021)Han, Rute, Wu, Ayers, and Polu]{han2021pact}
Han, J.~M., Rute, J., Wu, Y., Ayers, E.~W., and Polu, S.
\newblock Proof artifact co-training for theorem proving with language models.
\newblock \emph{arXiv preprint arXiv:2102.06203}, 2021.

\bibitem[Harrison(2009)]{harrison2009handbook}
Harrison, J.
\newblock \emph{Handbook of practical logic and automated reasoning}.
\newblock Cambridge University Press, 2009.

\bibitem[Jiang et~al.(2024)Jiang, Fonseca, and Cohen]{jiang2024leanreasoner}
Jiang, D., Fonseca, M., and Cohen, S.~B.
\newblock Leanreasoner: Boosting complex logical reasoning with lean.
\newblock \emph{North American Chapter of the Association for Computational
  Linguistics}, 2024.
\newblock \doi{10.48550/arxiv.2403.13312}.

\bibitem[Karunus \& Kovsharov(2024)Karunus and
  Kovsharov]{kovsharov2024paperproof}
Karunus, E. and Kovsharov, A.
\newblock Paperproof: A new proof interface for lean 4.
\newblock \url{https://github.com/Paper-Proof/paperproof}, 2024.
\newblock URL \url{https://github.com/Paper-Proof/paperproof}.
\newblock Accessed: 2025-04-30.

\bibitem[Kocsis \& Szepesv{\'a}ri(2006)Kocsis and
  Szepesv{\'a}ri]{kocsis2006uct}
Kocsis, L. and Szepesv{\'a}ri, C.
\newblock Bandit based monte-carlo planning.
\newblock In \emph{European conference on machine learning}, pp.\  282--293.
  Springer, 2006.

\bibitem[Lample et~al.(2022)Lample, Lacroix, Lachaux, Rodriguez, Hayat, Lavril,
  Ebner, and Martinet]{lample2022hypertree}
Lample, G., Lacroix, T., Lachaux, M.-A., Rodriguez, A., Hayat, A., Lavril, T.,
  Ebner, G., and Martinet, X.
\newblock Hypertree proof search for neural theorem proving.
\newblock \emph{Advances in neural information processing systems},
  35:\penalty0 26337--26349, 2022.

\bibitem[Limperg \& From(2023)Limperg and From]{aesop2023}
Limperg, J. and From, A.~H.
\newblock Aesop: White-box best-first proof search for lean.
\newblock In \emph{Proceedings of the 12th ACM SIGPLAN International Conference
  on Certified Programs and Proofs}, CPP 2023, pp.\  253–266, New York, NY,
  USA, 2023. Association for Computing Machinery.
\newblock ISBN 9798400700262.
\newblock \doi{10.1145/3573105.3575671}.
\newblock URL \url{https://doi.org/10.1145/3573105.3575671}.

\bibitem[Megill \& Wheeler(2019)Megill and Wheeler]{megill2019metamath}
Megill, N. and Wheeler, D.~A.
\newblock \emph{Metamath: a computer language for mathematical proofs}.
\newblock Lulu. com, 2019.

\bibitem[Morrison(2023)]{morrison2023lean-training-data}
Morrison, K.
\newblock lean-training-data: Tools for extracting training‑data from lean
  libraries.
\newblock \url{https://github.com/kim-em/lean-training-data}, 2023.
\newblock Accessed: 2025‑07‑01.

\bibitem[Moura \& Ullrich(2021)Moura and Ullrich]{moura2021lean}
Moura, L.~d. and Ullrich, S.
\newblock The {L}ean 4 theorem prover and programming language.
\newblock In \emph{Automated Deduction--CADE 28: 28th International Conference
  on Automated Deduction, Virtual Event, July 12--15, 2021, Proceedings 28},
  pp.\  625--635. Springer, 2021.

\bibitem[Polu \& Sutskever(2020)Polu and Sutskever]{polu2020gptf}
Polu, S. and Sutskever, I.
\newblock Generative language modeling for automated theorem proving.
\newblock \emph{arXiv preprint arXiv:2009.03393}, 2020.

\bibitem[Ren et~al.(2025)Ren, Shao, Song, Xin, Wang, Zhao, Zhang, Fu, Zhu,
  Yang, et~al.]{ren2025deepseek}
Ren, Z., Shao, Z., Song, J., Xin, H., Wang, H., Zhao, W., Zhang, L., Fu, Z.,
  Zhu, Q., Yang, D., et~al.
\newblock Deepseek-prover-v2: Advancing formal mathematical reasoning via
  reinforcement learning for subgoal decomposition.
\newblock \emph{arXiv preprint arXiv:2504.21801}, 2025.

\bibitem[Tooby-Smith(2024)]{tooby_smith2024formalization_physics}
Tooby-Smith, J.
\newblock Formalization of physics index notation in lean 4.
\newblock \emph{ArXiv}, abs/2411.07667, 2024.
\newblock URL \url{https://api.semanticscholar.org/CorpusID:273970152}.

\bibitem[Trinh et~al.(2024)Trinh, Wu, Le, He, and
  Luong]{trinh2024alpha_geometry}
Trinh, T.~H., Wu, Y., Le, Q.~V., He, H., and Luong, T.
\newblock Solving olympiad geometry without human demonstrations.
\newblock \emph{Nature}, 625\penalty0 (7995):\penalty0 476--482, 2024.

\bibitem[Tsoukalas et~al.(2024)Tsoukalas, Lee, Jennings, Xin, Ding, Jennings,
  Thakur, and Chaudhuri]{tsoukalas2024putnambench}
Tsoukalas, G., Lee, J., Jennings, J., Xin, J., Ding, M., Jennings, M., Thakur,
  A., and Chaudhuri, S.
\newblock Putnambench: Evaluating neural theorem-provers on the putnam
  mathematical competition.
\newblock \emph{arXiv preprint arXiv:2407.11214}, 2024.

\bibitem[van Doorn et~al.(2023)van Doorn, Massot, and
  Nash]{doorn2023sphere_eversion}
van Doorn, F., Massot, P., and Nash, O.
\newblock Formalising the h-principle and sphere eversion.
\newblock In \emph{Proceedings of the 12th ACM SIGPLAN International Conference
  on Certified Programs and Proofs}, CPP 2023, pp.\  121–134, New York, NY,
  USA, 2023. Association for Computing Machinery.
\newblock ISBN 9798400700262.
\newblock \doi{10.1145/3573105.3575688}.
\newblock URL \url{https://doi.org/10.1145/3573105.3575688}.

\bibitem[Wang et~al.(2025)Wang, Unsal, Lin, Baksys, Liu, Santos, Sung, Vinyes,
  Ying, Zhu, et~al.]{wang2025kimina}
Wang, H., Unsal, M., Lin, X., Baksys, M., Liu, J., Santos, M.~D., Sung, F.,
  Vinyes, M., Ying, Z., Zhu, Z., et~al.
\newblock Kimina-prover preview: Towards large formal reasoning models with
  reinforcement learning.
\newblock \emph{arXiv preprint arXiv:2504.11354}, 2025.

\bibitem[Whalen(2016)]{whalen2016holophrasm}
Whalen, D.
\newblock Holophrasm: a neural automated theorem prover for higher-order logic.
\newblock \emph{arXiv preprint arXiv:1608.02644}, 2016.

\bibitem[Wu et~al.(2024)Wu, Huang, Zhou, Ying, Wang, Lin, and
  Chen]{wu2024internlm2}
Wu, Z., Huang, S., Zhou, Z., Ying, H., Wang, J., Lin, D., and Chen, K.
\newblock Internlm2. 5-stepprover: Advancing automated theorem proving via
  expert iteration on large-scale lean problems.
\newblock \emph{arXiv preprint arXiv:2410.15700}, 2024.

\bibitem[Xin et~al.(2024{\natexlab{a}})Xin, Guo, Shao, Ren, Zhu, Liu, Ruan, Li,
  and Liang]{xin2024deepseek}
Xin, H., Guo, D., Shao, Z., Ren, Z., Zhu, Q., Liu, B., Ruan, C., Li, W., and
  Liang, X.
\newblock Deepseek-prover: Advancing theorem proving in llms through
  large-scale synthetic data.
\newblock \emph{arXiv preprint arXiv:2405.14333}, 2024{\natexlab{a}}.

\bibitem[Xin et~al.(2024{\natexlab{b}})Xin, Ren, Song, Shao, Zhao, Wang, Liu,
  Zhang, Lu, Du, et~al.]{xin2024deepseek-v1.5}
Xin, H., Ren, Z., Song, J., Shao, Z., Zhao, W., Wang, H., Liu, B., Zhang, L.,
  Lu, X., Du, Q., et~al.
\newblock Deepseek-prover-v1.5: Harnessing proof assistant feedback for
  reinforcement learning and monte-carlo tree search.
\newblock \emph{arXiv preprint arXiv:2408.08152}, 2024{\natexlab{b}}.

\bibitem[Yang et~al.(2023)Yang, Swope, Gu, Chalamala, Song, Yu, Godil, Prenger,
  and Anandkumar]{yang2023leandojo}
Yang, K., Swope, A., Gu, A., Chalamala, R., Song, P., Yu, S., Godil, S.,
  Prenger, R.~J., and Anandkumar, A.
\newblock Lean{D}ojo: Theorem proving with retrieval-augmented language models.
\newblock \emph{Advances in Neural Information Processing Systems},
  36:\penalty0 21573--21612, 2023.

\bibitem[Ying et~al.(2024{\natexlab{a}})Ying, Wu, Geng, Wang, Lin, and
  Chen]{ying2024lean}
Ying, H., Wu, Z., Geng, Y., Wang, J., Lin, D., and Chen, K.
\newblock Lean workbook: A large-scale lean problem set formalized from natural
  language math problems.
\newblock \emph{arXiv preprint arXiv:2406.03847}, 2024{\natexlab{a}}.

\bibitem[Ying et~al.(2024{\natexlab{b}})Ying, Zhang, Li, Zhou, Shao, Fei, Ma,
  Hong, Liu, Wang, et~al.]{ying2024internlm}
Ying, H., Zhang, S., Li, L., Zhou, Z., Shao, Y., Fei, Z., Ma, Y., Hong, J.,
  Liu, K., Wang, Z., et~al.
\newblock Internlm-math: Open math large language models toward verifiable
  reasoning.
\newblock \emph{arXiv preprint arXiv:2402.06332}, 2024{\natexlab{b}}.

\bibitem[Zheng et~al.(2021)Zheng, Han, and Polu]{zheng2021minif2f}
Zheng, K., Han, J.~M., and Polu, S.
\newblock Minif2f: a cross-system benchmark for formal olympiad-level
  mathematics.
\newblock \emph{arXiv preprint arXiv:2109.00110}, 2021.

\end{thebibliography}
